\documentclass[letterpaper, 10 pt, conference]{ieeeconf}  
\usepackage[dvipsnames]{xcolor}
\usepackage{soul}

\IEEEoverridecommandlockouts                              

\usepackage{balance}
\usepackage{cite}
\usepackage{graphicx} 
\usepackage{flushend}
\begin{document}

\title{\LARGE \bf
Robot Vulnerability and the Elicitation of User Empathy 
}

\author{Morten Roed Frederiksen$^{1}$ Katrin Fischer$^{2}$ Maja Matarić$^{3}$
  \thanks{$^{1}$Morten Roed Frederiksen {\tt\small mrof@itu.dk} is with the REAL lab at the Computer Science Department of the IT-University of Copenhagen. 
  $^{2}$Katrin Fischer {\tt\small katrinfi@usc.edu} is with the Communication Department at the University of Southern California, Los Angeles California.
  $^{3}$Maja Matarić {\tt\small mataric@usc.edu} is with the Computer Science Department at the University of Southern California, Los Angeles California.
  }
}
\maketitle
\begin{abstract}
This paper describes a between-subjects Amazon Mechanical Turk study (n = 220) that investigated how a robot’s affective narrative influences its ability to elicit empathy in human observers. We first conducted a pilot study to develop and validate the robot’s affective narratives.  Then, in the full study, the robot used one of three different affective narrative strategies (funny, sad, neutral) while becoming less functional at its shopping task over the course of the interaction. As the functionality of the robot degraded, participants were repeatedly asked if they were willing to help the robot. The results showed that conveying a sad narrative significantly influenced the participants’ willingness to help the robot throughout the interaction and determined whether participants felt empathetic toward the robot throughout the interaction. Furthermore, a higher amount of past experience with robots also increased the participants’ willingness to help the robot. This work suggests that affective narratives can be useful in short-term interactions that benefit from emotional connections between humans and robots.
\end{abstract}
\section{Introduction}

Robots that demonstrate vulnerability and elicit empathy have the potential to shape how humans communicate and to positively affect group dynamics \cite{traeger_vulnerable_2020}. For that reason, empathy has begun to be explored in socially assistive robotics (SAR) \cite{mataric_socially_2016}. Ideally, an empathetic bond is established on a foundation of genuine emotions that emerge from multiple interactions that may also increase attachment \cite{harvey_sex_2015}. However, not all SAR contexts involve repeated or long-term interactions. In those cases, it may be difficult for a robot to present itself in a way that establishes a connection with users. In short-term interactions, users rarely have time to establish a relationship with the robot and develop a sufficient representation of its behavior \cite{hegel_theory_2008}, posing a communication challenge for the robot \cite{stubbs_autonomy_2007}. 
This paper explores empathy elicitation by investigating interaction scenarios involving a robot that uses affective narratives to generate compassion while it is failing at its task. This work explores the relationship between the type of narrative conveyed by the robot (funny, sad, neutral) and the robot’s ability to elicit empathy in interactions with human observers. Specifically, the work explores whether a robot elicits more empathy when using affective narratives (funny, sad) than when not using affect (neutral) and how the affective narratives (funny vs. sad) compare in effectiveness in eliciting empathy. 
To pursue these research goals, we conducted an Amazon Mechanical Turk (AMT) \cite{mturk} experiment that presented human participants with a narrative conveyed by a robot that was performing a grocery shopping task while losing functionality over time. Participants were randomly assigned to three  groups that each were presented with one of three distinct narrative strategies: funny, sad, and neutral. The results show that presenting a sad narrative significantly increased the participants’ willingness to help the robot through all steps of the shopping task, and participants’ previous experiences with robots influenced their willingness to help the robot. Our findings provide novel insights about how narratives can be used to elicit user empathy in short-term interactions to swiftly catalyze an emotional connection between humans and robots.
\begin{figure}[h] \centering \includegraphics[width=0.489\textwidth]{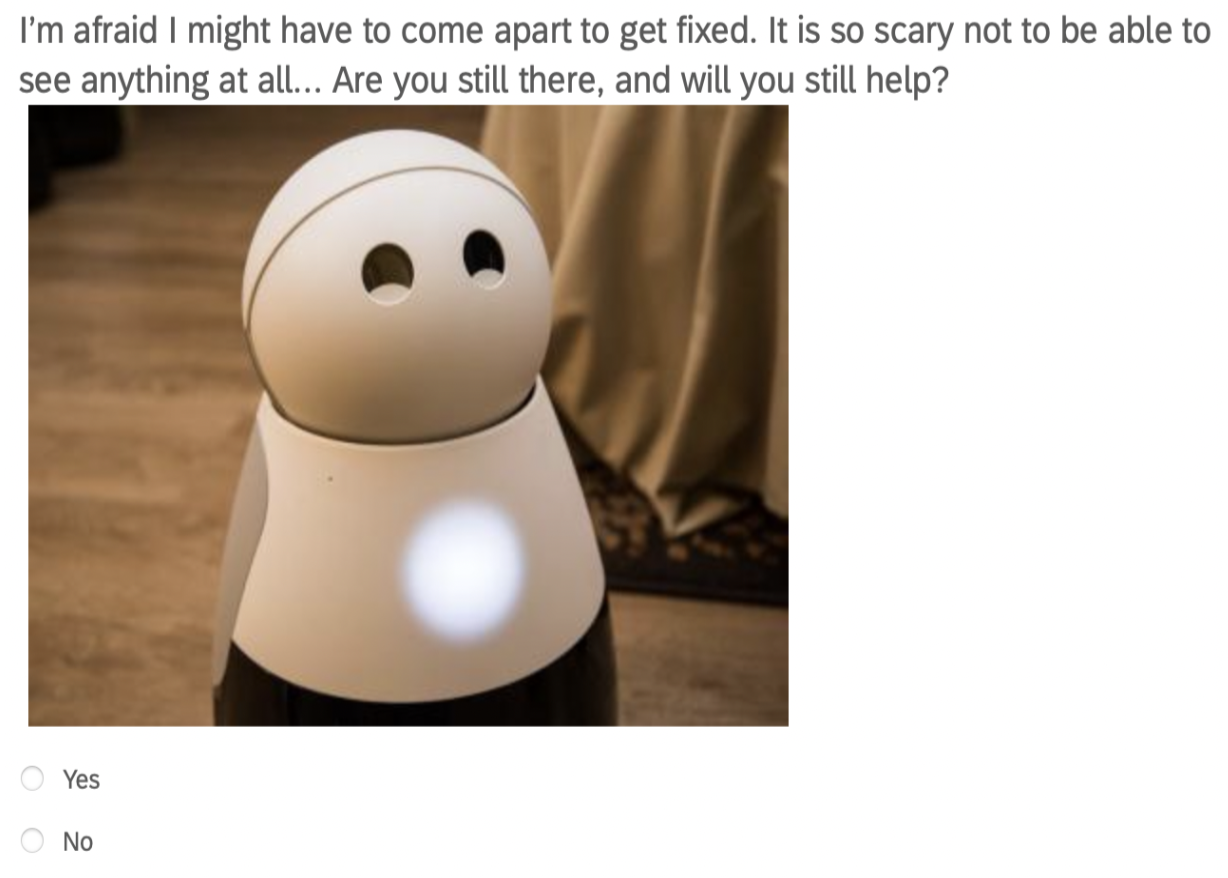} \caption{A single step of the AMT study.}
\label{kuri}
\end{figure}

This paper is organized as follows. Section II provides an overview of existing literature on empathy in HRI. Section III  describes the overall methodology used and details the user study design including the pilot study we conducted to validate the narrative strategies. Section IV provides an overview of the experiments, details the outcome measures used, and presents the main results. Section V discusses the findings and their implications, as well as potential future work directions. Section VI summarizes and concludes the paper.

\section{Previous approaches}
\subsection{Empathy in HRI}
Empathy refers to sharing the emotional state of another \cite{omdahl_empathy_2014} and the ability to identify with the other for the purpose of grasping their subjective experience \cite{covitz_salman_2009}. Although humans may be predisposed with the mental capacity to elicit empathy \cite{carr_neural_2003}, empathy can be trained as a skill \cite{riess_improving_2011}.  Empathy includes an affective component, i.e., sharing the emotional experience of another by having an affective (immediate) response, and a cognitive component, i.e., the ability to understand another’s perspective \cite{decety_functional_2004}. \cite{tapus_emulating_2007} outlined elements needed to enable robots to express empathy, including recognizing the other’s emotional state as well as expressing their own. 
Research into empathy in HRI  typically separates robots expressing empathy from robots eliciting empathy.  Previous work regarding the former area  found that positive empathic comments from a robot spectator can strengthen how the robot is perceived \cite{leite_influence_2013}. Participants who were assigned to the supportive version of a robot gave significantly higher ratings of companionship, reliable alliance, and self-validation. Similarly,  an experiment manipulating robot empathy in an elementary school setting showed that the empathetic behaviors of a robot had a positive impact on children’s perception of the robot \cite{leite_modelling_2012}. Another investigation found that robots communicating with empathy in their voice were perceived as being more engaged and able to convey empathy toward the patient \cite{james_artificial_2018}. Recent work \cite{tapus_emulating_2007} investigated the perceived effect of a robot displaying either cognitive or affective empathy and found that the robots that used immediate affective empathy were perceived as being more empathetic. 
In this paper, we focus on the second research area, robots eliciting empathy from humans. Past work shows that study participants reacted emotionally in response to videos of various treatments of robots, i.e., they felt more positive after viewing affectionate behavior toward robots and more negative (experienced emotional stress) after watching violent behavior toward robots \cite{james_artificial_2018}.  An examination of a robot’s influence in teams found that a vulnerable Nao robot increased the total talking time as well as relative talking time in the group conversation \cite{traeger_vulnerable_2020}. The effects of group dynamics and socially assistive robots were also investigated in \cite{short_modeling_2016, short_understanding_2017, short_robot_2017} in studies that demonstrated that robot mediators could positively impact group interactions, including increasing cohesion and reducing group members’ selfish behavior.  A subordinate robot may also encourage empathy \cite{deng_embodiment_2019}; a study examining people’s reactions when witnessing robot mistreatment  showed that participants were more likely to intervene when bystander robots exhibited an anthropomorphic sad response \cite{connolly_prompting_2020}.
 
\subsection{Empathy elicited by the narrative}

Previous studies have explored how storytelling by a robot can help to elicit user empathy \cite{mathur_modeling_2021}. Since empathy involves identification with another to comprehend their subjective experience \cite{covitz_salman_2009}, it is important to study how narratives can serve as means of evoking empathy. Theories of narrative empathy aim to establish how people relate to believable worlds in the literature; they discuss how narrative strategies evoke empathetic responses through considerations of narrative situation and perspective \cite{keen_suzanne_theory_nodate}. Studies show that even when empathy is successfully elicited, it is not static and must be maintained or it declines over time \cite{hojat_devil_2009}. There is evidence that the first-person perspective of a narrative conveyed by a robot influences the level of empathy elicited by human observers and increases the trustworthiness of the robot \cite{spitale_socially_2018}.  There are also narrative strengths in employing serial repetition of narratives set in a stable story world \cite{warhol_having_2003}.
\section{Method}
This work explored the following research questions:

\textbf{RQ1}. How do different narrative strategies (funny, sad, neutral) differ in their ability to elicit participants’ willingness to help? 

\textbf{RQ2}.  How do funny and sad narratives compare to the neutral narrative in their ability to generate empathy for a robot?

We designated the drop-off point as a key outcome measure in this work.  The drop-off point is defined as the first narrative step at which participants respond “No” when asked to help the robot.
We defined the following hypotheses:

\textbf{H1}. A robot using an affective narrative strategy (happy or sad) will receive help longer, i.e., will have a later drop-off point,  than a robot using a neutral narrative strategy. 

\textbf{H2}. The average number of “Yes” responses to the robot's request for help (including participants who dropped off and then returned) will be significantly higher in the sad and funny narrative strategies than in the neutral strategy.

\textbf{H3}. The probability of reaching the final (24th) step of the interaction while continuously helping the robot will be significantly higher for participants exposed to the sad and funny narrative strategies.

\textbf{H4}. The funny and sad affective narrative strategies will be significant predictors of self-reported empathy. 

\textbf{H5}. Familiarity with robots will be a significant predictor of elicited empathy as defined by a willingness to help the robot.  Specifically, participants more familiar with robots will be more likely to help the robot.

\begin{figure}[h] \centering \includegraphics[width=0.3\textwidth]{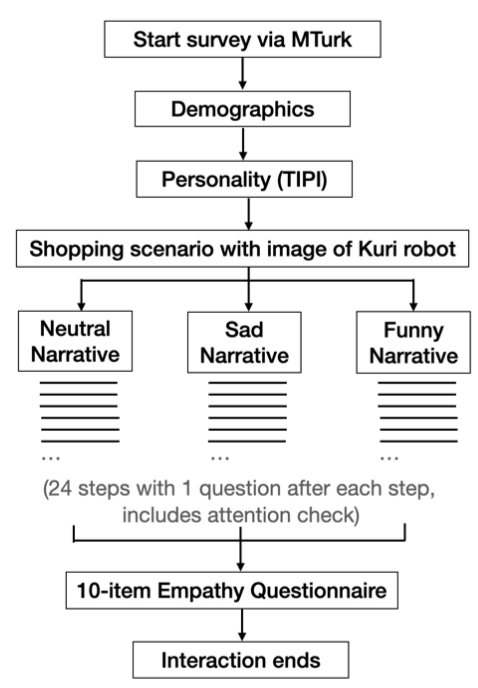} \caption{Flowchart of  administered study questionnaires.}
\label{flowchart}
\end{figure}
\subsection{Study Design}
The study used a between-subjects design with the robot’s affective narrative strategy as the independent variable. Each narrative contained a specific emotional direction (happy, sad, or neutral) and the robot employed that direction to establish an emotional connection with the participant. Study participants were shown an image of a Kuri robot \cite{kuri} and a short description of the situation, then they engaged with the robot in a total of 24 narrative steps.  The image was consistent and generic across the narrative steps to keep the participants' focus on the narrative and refrain from introducing other variables.

\subsection{Participants \& Measures}
Power for any given study is recommended to be at least 0.8 \cite{cohen_statistical_1988}. Power analysis with G*Power OS application version 3.1.9.3 for fixed effects, omnibus, one-way ANOVA with 3 groups and effect size of $f^2$ = 0.25 suggested a total sample size of at least 159 participants for this study. Accordingly, a total of 305 participants were recruited via Amazon Mechanical Turk \cite{mturk}. 85 people failed the attention check and their answers were discarded, resulting in 220 participants who were randomly assigned to one of the three affective narrative strategies. 
Demographic questions assessed participants’ age (in years) and familiarity with robots (on a 5-point Likert scale ranging from “not familiar at all” to “extremely familiar”). Furthermore, the experiment included a 10-item personality questionnaire (TIPI) \cite{gosling_very_2003} to investigate whether personality traits affected participants' empathic responses. A post-task questionnaire also measured participants’ self-reported empathetic feelings toward the robot. The survey items were inspired by and adapted from Davis’ Interpersonal Reactivity Index (IRI) \cite{iri}, Connolly et al. \cite{connolly_prompting_2020}, and Rosenthal-von der Pütten et al. \cite{rosenthal-von_der_putten_investigations_2014}. All were answered on a 7-point Likert scale ranging from Strongly Disagree to Strongly Agree (note that (-) indicates reverse-coded items): 1. “When I saw the problems the robot was having, I felt sad”, 2. “I found it difficult to empathize with the robot” (-), 3. “I felt protective of the robot”, 4. “I didn’t feel very sorry for the robot when it was having problems” (-), 5. “I felt helpless watching the robot deal with the technical difficulties”, 6. “When I saw the robot malfunctioning, I remained calm” (-), 7. “I felt for the robot”, 8. “The robot’s misfortune did not disturb me a great deal” (-), 9. “I sympathized with the robot”, 10. “Seeing the robot in this situation did not affect me” (-). The aggregate of these items formed the empathy index.

\subsection{Procedure}
The participants were randomly assigned to one of the three affective narrative strategies: happy, sad, or neutral. The narrative in each strategy consisted of 24 steps of interaction between a Kuri robot \cite{kuri} and the study participant. Each step was presented as an image of the robot and its utterance shown as a text caption above the image, as shown in Figure \ref{kuri}, above.  The robot attempted to complete a mock grocery shopping task. As it did so, the robot had problems at each step (with its vision system, its mobility, etc.), and requested help from the participant. At each step, the participant was given the choice (via a Yes/No question under the robot’s caption and image) of helping the robot.   
The affective narrative strategies were designed to elicit as much participant empathy as possible by meeting the following criteria: a) offer the best possibility for a participant to identify with the situation \cite{covitz_salman_2009}: the robot’s problems were relatable as they regarded seeing and moving; b), explore the decaying effect of empathy: the 24 steps involved varying stimuli in the form of different events, inspired by \cite{connolly_prompting_2020}; c) induce more empathy over time, per \cite{gosling_very_2003}: the robot’s functionality decreased continually through the interaction; and e) leverage compelling storytelling as a means of inducing empathy \cite{zak_why_2015}: the 24-step interaction was set up so it appeared like a developing story. These criteria were specifically designed to improve written narratives and may not be applicable for human-robot interactions. They functioned mainly as inspiration for designing our narratives.

\section{Results}

\subsection{Pilot Study}
We conducted a pilot study to validate the intended effect of each narrative. We composed a 14-step narrative for each affective strategy (funny, sad, neutral). A total of 308 U.S. AMT participants were recruited.  At each step a different error was presented by the robot and the participant was offered the choice of helping or refusing to help the robot. Following each step, the participant was asked to rate the robot’s phrase as funny, sad, or neither, and indicated with a Yes/No whether they felt sorry for the robot. If they answered No about feeling sorry for the robot, they were asked to clarify 
what they felt (write-in box). If a participant chose not to help the robot, they exited the study at that point of the narrative and did not complete the remaining steps.

\subsection{Pilot Study Results}
\begin{figure}[h] \centering \includegraphics[width=0.5\textwidth]{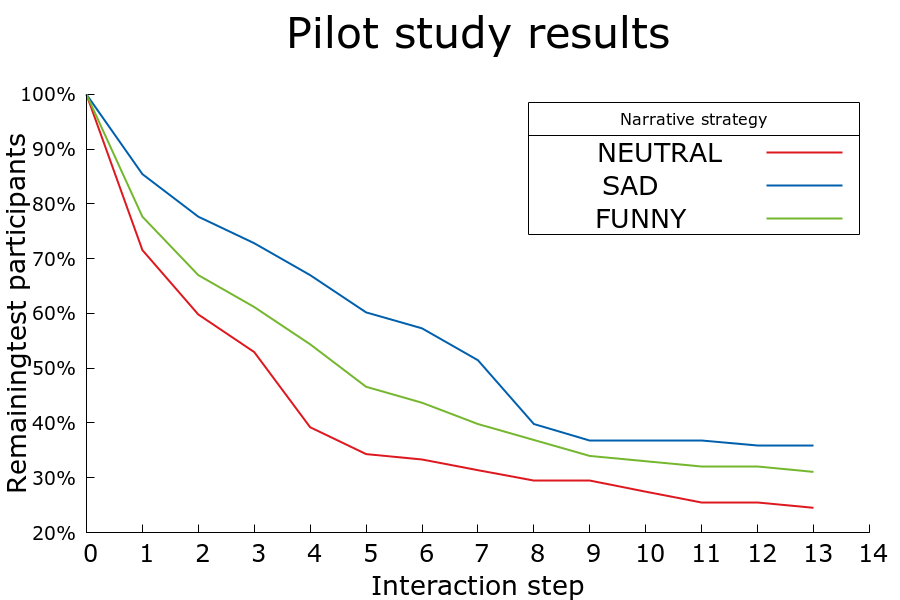} \caption{Plot of the drop-off rate of participants over the course of the 14-step pilot study for the three conditions. The drop-off is plotted as the percentage of remaining participants at each interaction step, defined as the share of participants who have agreed to help the robot at any previous step.}
\label{pilot}
\end{figure}

Initial events are important in conveying a compelling narrative: There was a noticeable drop-off after the first 4-6 steps, with about 50\% of the participants having exited (i.e., stopped helping the robot) in the funny and neutral affective narrative strategies. More specifically, at step 5, only 40 participants (out of 102) remained in the neutral narrative strategy, while 56 (out of 103) participants remained in the funny strategy and 69 (out of 103) in the sad strategy. The drop-off rate increase was most pronounced in the neutral strategy throughout the narrative. The drop-off in the sad narrative tapered off more slowly and reached 50\% between steps 8 and 9, when it dropped from 53 to 41 participants  (out of 103), as shown in Figure \ref{pilot}. 
Empathy is key for persuading participants to help: On average, almost all participants (95\%) who reported feeling sorry for the robot helped the robot and stayed for the next step, regardless of the affective narrative strategy. In the sad narrative strategy, 96\% of participants who said they felt sorry agreed to help the robot (averaged over all 14 steps). Feeling sorry therefore appears to be a good indication of who will help the robot. Conversely, of all participants who helped, 87\% said they were feeling sorry, while 13\% reported other motivations (e.g., they felt neutral, curious). 
The pilot study helped us to identify the following necessary changes for the full study:
The clarity of some narrative content: we eliminated content that was misinterpreted by participants.
The amount of variability in robot errors presented with each event: We originally used 6 different error types: Speaker malfunction, Microphone noise, navigational error, vision system failure, Logical error, General System malfunction. We changed this to use variations around a vision failure and a single display of a mobility failure. The vision failure type was selected as being a highly plausible error. The total length of the interaction: we added 10 steps for a total of a 24-step full study, and instead of terminating the interaction at the point when participants chose not to help the robot, we extended the interaction to the full 24 steps to gain more information. More than 25\% of participants from the pilot completed the full set of steps. We were interested in knowing if this tendency would persist if we increased the length of the interaction. Having participants complete the full set of steps also allowed us to gain information about whether participants who had previously denied help to the robot would later regain empathy and agree to help the robot. We chose to use willingness to aid the robot as a measure of elicited empathy in the full study because the pilot study confirmed that  87\% of the participants who decided to help the robot were motivated by feeling sorry for the robot and, of those who felt sorry for the robot, 95\% helped the robot.
\subsection{Full Study Design}
The full study design was identical to the pilot study, except for the above-listed improvements: increased interaction length (24 steps), no early termination, single robot error per step and single error type throughout the interaction. Measures of personality and self-reported empathy were added.
\subsection{Study Results}
220 AMT participants completed the full study and passed the attention check. 153 identified as male and 67 as female (none selected “Other”). Participants’ ages ranged from 21 to 70 years old ($M = 34.7$, $SD = 10.5$) and their college education (post high school) ranged from 0 to 11 years ($M = 3.8$, $SD = 1.98$)  Participants’ familiarity with robots was diverse, between 1 (not familiar at all) to 5 (extremely familiar) with a mean of 3.7 ($SD = 1.0$). Aggregates of personality (TIPI) questions indicated that participants represented the whole spectrum (range 1-7) of introversion/extroversion ($M = 4.0$, $SD = 1.4$),  agreeableness ($M = 4.7$, $SD = 1.3$) and conscientiousness ($M = 4.8$, $SD = 1.3$). Emotional stability ($M = 4.6$, $SD = 1.4$) and openness to experience ($M = 4.6$, $SD = 1.3$) responses on the aggregated index ranged from 1.5 to 7. Participants answered all 10 empathy questionnaire items as described in Section III (B). Exploratory factor analysis showed that both a unidimensional scale and a two-factor solution would be viable. The unidimensional empathy scale had high factor loadings for all items except for item 5 (“I felt helpless watching the robot deal with the technical difficulties”) and 6 (reverse-coded, “When I saw the robot malfunctioning, I remained calm”). The resulting empathy index was created with the eight viable items (Cronbach’s alpha = 0.83). The two-factor solution produced a factor with all reverse-coded items (except item 6) in one factor and all positively worded empathy statements in another, which was also considered for analysis. 

\begin{figure}[h] \centering \includegraphics[width=0.5\textwidth]{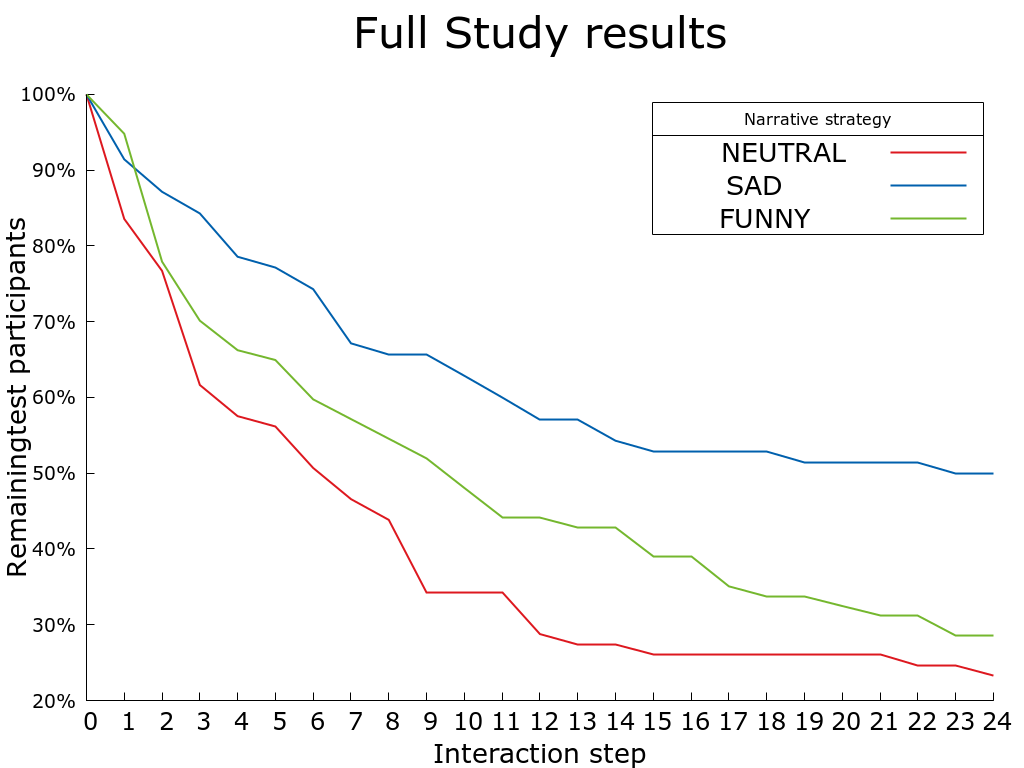} \caption{Plot of the drop-off rate of participants over the course of the 24-step pilot study for the three conditions. The drop-off is plotted as the percentage of remaining participants at each interaction step, defined as the share of participants who have agreed to help the robot at any previous step. This figure only depicts the participants’ initial refusal to aid the robot and it does not reflect the data for those who re-engaged with the robot.}
\label{main_study}
\end{figure}

\textit{A sad narrative has a significant impact on the participants’ sustained streak of willingness to keep helping the robot:} A one-way between-subjects ANOVA was conducted to compare the effect of the narrative (funny, sad, neutral) on the drop-off point (step 1 - 24). Assumptions for parametric tests were met (Levene’s test was not significant). Results showed that the effect of the type of narrative on the dropoff point, i.e. how long participants chose to keep helping the robot without disagreeing, was significant, $F(2, 217) = 6.58$, $p = 0.0017$. Post hoc comparisons using the Tukey HSD confirmed that the sad (but not the funny) narrative produced a significant difference to the control strategy, i.e. the neutral narrative. Therefore, \textbf{H1} is partially supported. Affective narrative strategies did not have an impact on the total number of “Yes” responses. While the sad narrative initially helped to keep participants helping the robot, it did not have a significant impact when taking into account participants who had dropped off and later re-engaged. \textbf{H2} is therefore not supported. Figure \ref{main_study} visualizes the drop-off of test participants and highlights the significantly less steep drop-off for the sad narrative (shown in blue) in comparison with the neutral strategy (shown in red). Of the participants who chose not to help the robot but then decided to re-engage and help the robot again, the average number of steps it took for them to re-engage was also smallest with the sad strategy (2.26) followed by the funny strategy (2.34), and then the neutral strategy (2.40); the difference was not statistically significant.  
\textit{Presenting a  sad narrative significantly impacts the participants’ willingness to keep interacting with the robot until the  final interaction step:} A logistic regression with the binary outcome variable of whether participants are still willing to help the robot at narrative step 24 confirmed that the sad narrative significantly influenced the outcome, $b = 1.192$, $z =  3.26$, $p = 0.0011$. Being exposed to the sad narrative strategy meant that the probability of continually helping the robot until step 24 was 3.29 times higher (an increase of 329\%) than being exposed to the neutral strategy. Being exposed to the funny strategy did not significantly predict the outcome, although the probability of continually helping was 1.31 times higher (131\%) than for the neutral strategy. \textbf{H3} is therefore partially supported.  Affective strategies did not have an effect on self-reported empathy; \textbf{H4} is not supported. 
The results of a multiple linear regression model with the empathy index as the dependent variable indicated that there was a collective significant effect between the five personality traits and self-reported empathy ($F(5, 214) = 10.8$, $p < .01$, Adjusted $R^2 = 0.182$). The individual predictors were examined further and indicated that agreeableness ($b = 0.2289$, $SE = 0.0739$, $t(214) = 3.10$, $p = 0.0022$), but no other tested personality trait, was a significant predictor of self-reported empathy.

\textit{Empathy may be a learnable skill:} A linear regression model with familiarity as the predictor and helping the robot as outcome variable showed that familiarity with robots significantly predicted how many times participants agreed to help, i.e. responded “Yes” to all 24 steps ($b = 1.322$, $SE = 0.387$, $t(218) = 3.42$, $p = 0.00076$. Familiarity also significantly predicted self-reported empathy for the positively worded statements in Factor 1 ($b = 0.4221$, $SE = 0.0822$, $t(218) = 5.135$, $p = 6.25e-07$). This indicates that past exposure to robots can influence how people continue to interact with them and may increase the potential for them to invest emotionally in an interaction with a robot. \textbf{H5} is therefore supported.

\section{Discussion}
The immediate and steep drop-off in the participants’ willingness to help the robot indicates that the actions taken by the robot in the initial moments of an interaction may be highly influential in eliciting empathy from the participants. First impressions may influence the level of elicited trust \cite{xu_impact_2018}. In our pilot study, 90\% of the participants had not dropped off after the first step.  As the initial moments of an interaction present a key opportunity to convey affective information it may be argued that it is beneficial to have robots skip any introductory comments and immediately engage users with a compelling narrative. In our full study, the robot used no introductory comments and instead started conveying its problems and the plea for help. The results show an impact on the drop-off rate from steps 1-5 for all narratives with the largest difference registered for the sad narrative that retains 29\% more participants at step 5 for the full study compared to the pilot study. Presenting a more detailed background story before initiating the interaction could potentially have a positive effect on the participants perception of the robot, but doing so could also distract the participants from the initial moment when the robot has their maximum attention. We chose to use this crucial initial moment to convey affective narratives. Our results indicate that this approach had the intended effect.

\textit{Matching the narrative strategy to the intended task:} The pilot and the full study showed that participants stayed longer to help the robot when presented with the sad narrative. This result highlights the importance of the robot’s behavior and narrative strategy as it attempts to elicit empathy. Further, this result suggests that there is a benefit to portraying events using a narrative that emphasizes anthropomorphic interpretations of the robot when the aim is to elicit human empathy. Although the other narrative strategies (happy, neutral) were less effective than the sad narrative, both of these strategies managed to retain above 35\% of the participants at step 5 and they also both managed to persuade participants to aid them up until the final step. Therefore, those strategies may be usable in a different interaction context. For instance, the effectiveness of both the sad and funny narratives may be culturally dependent. This could mean that a neutral narrative robot may be overall less effective at evoking empathy in populations like the AMT participants in our study, but could be desirably culturally agnostic and thus able to function across different cultures. Similar observations were outlined in \cite{barolli_experimental_2020}.
This work investigated both the overall inclination toward helping the robot through the total recorded positive (Yes) answers and the sustained empathy toward the robot. The overall number of Yes answers per user from each narrative group was not significantly affected by the type of narrative. However, participants’ drop-off points were significantly postponed when using the sad narrative. A possible explanation might be that the overall number of Yes answers can be seen as reflecting the participants’ overall level of positivity toward robots. This underlying attitude may not necessarily be altered during a short interaction. Conversely, the sustained willingness to aid the robot seems far more sensitive to and influenced by the immediate events of the interaction. This may explain why the drop-off points were postponed by using the sad and funny narrative strategies compared with the neutral strategy. An effect on the sustained willingness to help the robot could also prove more useful in real world domains, as a user would most likely leave the conversation after once having refused to help the robot. 

\textit{Persistent robot behaviors may be beneficial:} This work attempted to investigate if using either of the affective narrative strategies (happy, sad) proved more successful in re-engaging participants who chose not to help the robot. The results show that all three narrative strategies were successful in re-engaging the participants in the interaction, as indicated by the average streak of No answers for all three narratives being between 2 and 3. This suggests that using persistent behaviors could have an effect on accomplishing tasks if they involve gaining human help. 
The average number of steps it took before participants re-engaged was a lower for the sad narrative strategy, possibly indicating a small advantage of using such a strategy if a given task requires a robot to re-engage humans in an interaction.

Agreeableness was found to be a significant predictor of self-reported empathy. Personality traits are generally considered to be relatively stable internal dispositions \cite{butrus_personality_2013}. Agreeableness is one of the five widely accepted personality dimensions \cite{mccrae_validation_1987}, responsible for prosocial behavior, i.e., behavior that benefits or helps others and is thought to be correlated with empathy itself \cite{mooradian_dispositional_2011}. Previous work found significant relations between personality and self-reported empathy \cite{graziano_agreeableness_2007} hence our finding that agreeableness is a significant predictor of self-reported empathy concurs with existing theories and extends them to include empathetic behavior toward robots.

Future studies could examine real-world human-robot interactions to generalize findings beyond the AMT context. Research questions for further investigation include the context in which these affective narrative strategies may be used. For instance, although the sad narrative strategy performed best in the shopping context we created, it may be the case that a funny narrative strategy would work better in other contexts or with a different user segment. For instance, since children often use anthropomorphic interpretations of inanimate objects \cite{severson_imagining_2018}, they may be more engaged with a funny narrative robot.  This work aims  to stimulate further investigations into the match between social context and the effectiveness of narrative strategies, and suggests comparing the effect found here with more or less anthropomorphic agents.

Although this work has suggested that the choice of narrative strategy has a large impact on the human-robot interaction, there may be further contextual variables that influence the amount of empathy elicited by the robot in a scenario. In our studies, the neutral strategy was created with the aim to be devoid of emotions. Even though its dialogue consisted solely of status messages, it could still elicit empathy from some participants. This indicates that the presented context itself may also impact the level of elicited empathy and presents future research opportunities into the synergies between narrative strategies and the interaction context.
\section{Conclusion}
This work explored the impact of different affective narrative strategies on eliciting empathy in short-term human-robot AMT interactions. A pilot study ($n=308$) and a full study ($n=220$) investigated  hypotheses about the efficacy of happy, sad, and neutral narrative strategies on eliciting empathy in participants and sustaining their willingness to keep providing help to the robot whose functionality was declining over subsequent interaction steps.  The results show that using a sad narrative strategy had a significant impact on sustaining participants’ willingness to help the robot. However, neither the sad nor funny narrative strategies produced a higher number of total positive answers toward aiding the robot per interaction. Finally, the study participants’ familiarity with robots was found to be a predictor of their willingness to help the robot during the interaction which was confirmed with a significant result. 
Eliciting empathy from humans is a highly complex process. Even current neurocognitive research have yet to grasp the physical relationship between affective empathy and cognitive empathy \cite{yu_dual_2018}. This paper shows a way to elicit empathy by using consistent behaviors and expressing robot struggles in a 1st person narrative. The project showed that the amount of elicited empathy in an interaction can be influenced by the narrative strategy and that it may be beneficial to emphasize sadness in the narrative when the aim is to sustain user interest and willingness to help a robot. We recommend further investigations into how real-world interactions may influence the reported AMT results and how contextual changes and different user segments may alter the impact of different narrative strategies.

\begin{center}
    \textbf{Acknowledgements} 
\end{center}
We thank Kourtney Chima at the University of Southern California for helping to set up the pilot study. 

\begin{center}
    \textbf{Citation diversity statement.} 
\end{center}
Recent work in several fields of science has identified a bias in citation practices such that papers from women and other minority scholars are under-cited relative to the number of such papers in the field. To heighten the awareness of this problem, we state the male/female distribution and attempt to include references such that they count towards decreasing the bias. Of the included references, 44\% have been published by a Female/Male team with a female lead writer, 12\% by Male/Female with a male lead writer, 16\% by a solely female team or individual, and 28\% by a solely male team or individual writer.

\bibliography{bibliography}

\balance
\end{document}